\documentclass{article}

\usepackage{hyperref}

\usepackage[accepted]{icml2019}

\usepackage{graphics}      

\usepackage{savesym}
\usepackage{amsmath}

\usepackage{color}

\usepackage{microtype}        
\usepackage{ccicons}          

\usepackage{relsize} 
\usepackage{mathtools,xparse}
\usepackage[table]{xcolor}
\usepackage{verbatim}
\usepackage{comment}
\usepackage{url}
\usepackage{soul}
\usepackage{multicol} 
\usepackage{multirow}
\usepackage{colortbl} 
\usepackage{amssymb}
\usepackage{wasysym}
\makeatletter
\def\oversortoftilde#1{\mathop{\vbox{\m@th\ialign{##\crcr\noalign{\kern3\p@}%
				\sortoftildefill\crcr\noalign{\kern3\p@\nointerlineskip}%
				$\hfil\displaystyle{#1}\hfil$\crcr}}}\limits}

\def\sortoftildefill{$\m@th \setbox\z@\hbox{$\braceld$}%
	\braceld\leaders\vrule \@height\ht\z@ \@depth\z@\hfill\braceru$}

\makeatother

\usepackage{tikz-cd}

\usepackage{tabulary}
\usepackage{booktabs}

\usepackage[english]{babel}
\usepackage{enumitem}

\newtheorem{lemma}{Lemma}

\newtheorem{definition}{Definition}[section]

\usepackage{array}
\usepackage{bytefield}

\usepackage[labelformat=simple,font={normal}, labelsep=period,singlelinecheck=on]{caption}
\usepackage[labelformat=simple,font={normal}, labelsep=period,singlelinecheck=on]{subcaption}

\usepackage{cleveref}
\crefname{section}{§}{§§}
\Crefname{section}{§}{§§}

\usepackage{slashbox}
\usepackage{multirow}
\usepackage{xcolor}

\usepackage{dsfont}
\usepackage{transparent}
\usepackage{import}
\usepackage{calrsfs}
\DeclareMathAlphabet{\pazocal}{OMS}{zplm}{m}{n}

\DeclareMathAlphabet\mathbfcal{OMS}{cmsy}{b}{n}

\usepackage{amsfonts}

\newcolumntype{C}{c<{\kern\tabcolsep}@{}}

\usepackage{tikz}
\definecolor{mygreen}{RGB}{0,176,80}
\definecolor{myred}{RGB}{102,0,0}
\definecolor{myblue}{RGB}{0,0,102}
\definecolor{myblue2}{RGB}{0,51,102}
\definecolor{myblue3}{RGB}{0,76,153}   
\definecolor{myblue4}{RGB}{48, 144, 199}  

\usetikzlibrary{decorations.pathreplacing}
\usetikzlibrary{bending}
\usepackage{lipsum}

\usetikzlibrary{positioning}
\tikzset{main node/.style={circle,fill=blue!20,draw,minimum size=1cm,inner sep=0pt},
}

\usepackage{etoolbox}
\patchcmd{\footnotemark}{\stepcounter{footnote}}{\refstepcounter{footnote}}{}{}

\usepackage{tabulary}
\newcolumntype{K}[1]{>{\centering\arraybackslash}p{#1}}
\usepackage{makecell}

\usepackage{tikzit}
\usepackage{pspicture}
\usepackage{pstricks}
\usepackage{graphicx}
\usepackage{amssymb}
\usepackage{wasysym}
\usetikzlibrary{positioning}
\usetikzlibrary{shapes,arrows}
\usetikzlibrary{calc}
\usetikzlibrary{decorations.pathreplacing}

\makeatletter
\pgfkeys{/pgf/.cd,
	parallelepiped offset x/.initial=5mm,
	parallelepiped offset y/.initial=5mm
}
\pgfdeclareshape{parallelepiped}
{
	\inheritsavedanchors[from=rectangle] 
	\inheritanchorborder[from=rectangle]
	\inheritanchor[from=rectangle]{north}
	\inheritanchor[from=rectangle]{north west}
	\inheritanchor[from=rectangle]{north east}
	\inheritanchor[from=rectangle]{center}
	\inheritanchor[from=rectangle]{west}
	\inheritanchor[from=rectangle]{east}
	\inheritanchor[from=rectangle]{mid}
	\inheritanchor[from=rectangle]{mid west}
	\inheritanchor[from=rectangle]{mid east}
	\inheritanchor[from=rectangle]{base}
	\inheritanchor[from=rectangle]{base west}
	\inheritanchor[from=rectangle]{base east}
	\inheritanchor[from=rectangle]{south}
	\inheritanchor[from=rectangle]{south west}
	\inheritanchor[from=rectangle]{south east}
	\backgroundpath{
		\southwest \pgf@xa=\pgf@x \pgf@ya=\pgf@y
		\northeast \pgf@xb=\pgf@x \pgf@yb=\pgf@y
		\pgfmathsetlength\pgfutil@tempdima{\pgfkeysvalueof{/pgf/parallelepiped offset x}}
		\pgfmathsetlength\pgfutil@tempdimb{\pgfkeysvalueof{/pgf/parallelepiped offset y}}
		\def\ppd@offset{\pgfpoint{\pgfutil@tempdima}{\pgfutil@tempdimb}}
		\pgfpathmoveto{\pgfqpoint{\pgf@xa}{\pgf@ya}}
		\pgfpathlineto{\pgfqpoint{\pgf@xb}{\pgf@ya}}
		\pgfpathlineto{\pgfqpoint{\pgf@xb}{\pgf@yb}}
		\pgfpathlineto{\pgfqpoint{\pgf@xa}{\pgf@yb}}
		\pgfpathclose
		\pgfpathmoveto{\pgfqpoint{\pgf@xb}{\pgf@ya}}
		\pgfpathlineto{\pgfpointadd{\pgfpoint{\pgf@xb}{\pgf@ya}}{\ppd@offset}}
		\pgfpathlineto{\pgfpointadd{\pgfpoint{\pgf@xb}{\pgf@yb}}{\ppd@offset}}
		\pgfpathlineto{\pgfpointadd{\pgfpoint{\pgf@xa}{\pgf@yb}}{\ppd@offset}}
		\pgfpathlineto{\pgfqpoint{\pgf@xa}{\pgf@yb}}
		\pgfpathmoveto{\pgfqpoint{\pgf@xb}{\pgf@yb}}
		\pgfpathlineto{\pgfpointadd{\pgfpoint{\pgf@xb}{\pgf@yb}}{\ppd@offset}}
	}
}
\makeatother

\newsavebox\CBox

\icmltitlerunning{Submission and Formatting Instructions for ICML 2019}

\begin{document}

\twocolumn[
\icmltitle{A Fast-Optimal Guaranteed Algorithm For Learning Sub-Interval Relationships in Time Series}




\begin{icmlauthorlist}
\icmlauthor{Saurabh Agrawal}{umn}
\icmlauthor{Saurabh Verma}{umn}
\icmlauthor{Anuj Karpatne}{umn}
\icmlauthor{Stefan Liess}{umn-es}
\icmlauthor{Snigdhansu Chatterjee}{umn-stat}
\icmlauthor{Vipin Kumar}{umn}
\end{icmlauthorlist}

\icmlaffiliation{umn}{Department of Computer Science, University of Minnesota}
\icmlaffiliation{umn-es}{Department of Earth Sciences, University of Minnesota}
\icmlaffiliation{umn-stat}{School of Statistics, University of Minnesota}

\icmlcorrespondingauthor{Saurabh Agrawal}{agraw066@umn.edu}

\icmlkeywords{Machine Learning, ICML}

\vskip 0.3in
]



\printAffiliationsAndNotice{} 

\begin{abstract}
Traditional approaches  focus on finding relationships between two entire time series, however, many interesting relationships exist in small sub-intervals of time and remain  feeble during other sub-intervals. We define the notion of a sub-interval relationship (SIR) to capture such interactions   that are prominent only in certain sub-intervals of time. To that end, we propose a fast-optimal guaranteed algorithm to find most interesting SIR relationship in a pair of time series. Lastly, we demonstrate the utility of our method in climate science domain based on a real-world dataset  along with its scalability scope and obtain useful domain insights. 
 
\end{abstract}

\section{Introduction}
Mining relationships in time series data is of immense interest to several disciplines such as neuroscience, climate science, and transportation. For example, in climate science, relationships are studied between time series of physical variables such as Sea Level Pressure, temperature, etc., observed at different locations on the globe. Such relationships, commonly known as `teleconnections' capture the underlying processes of the Earth's climate system \cite{kawale2013graph}. Similarly, in neuroscience, relationships are studied between activities recorded at different regions of the brain over time \cite{atluri2016brain,atluri2015connectivity}. Studying such relationships can help us improve our understanding of real-world systems, which in turn, could play a crucial role in devising solutions for   problems such as mental disorders or climate change.


\begin{figure*}[h]
	\centering
	\begin{subfigure}[t]{0.22\textwidth}
		\vspace{-12em}
		\includegraphics[scale=0.29]{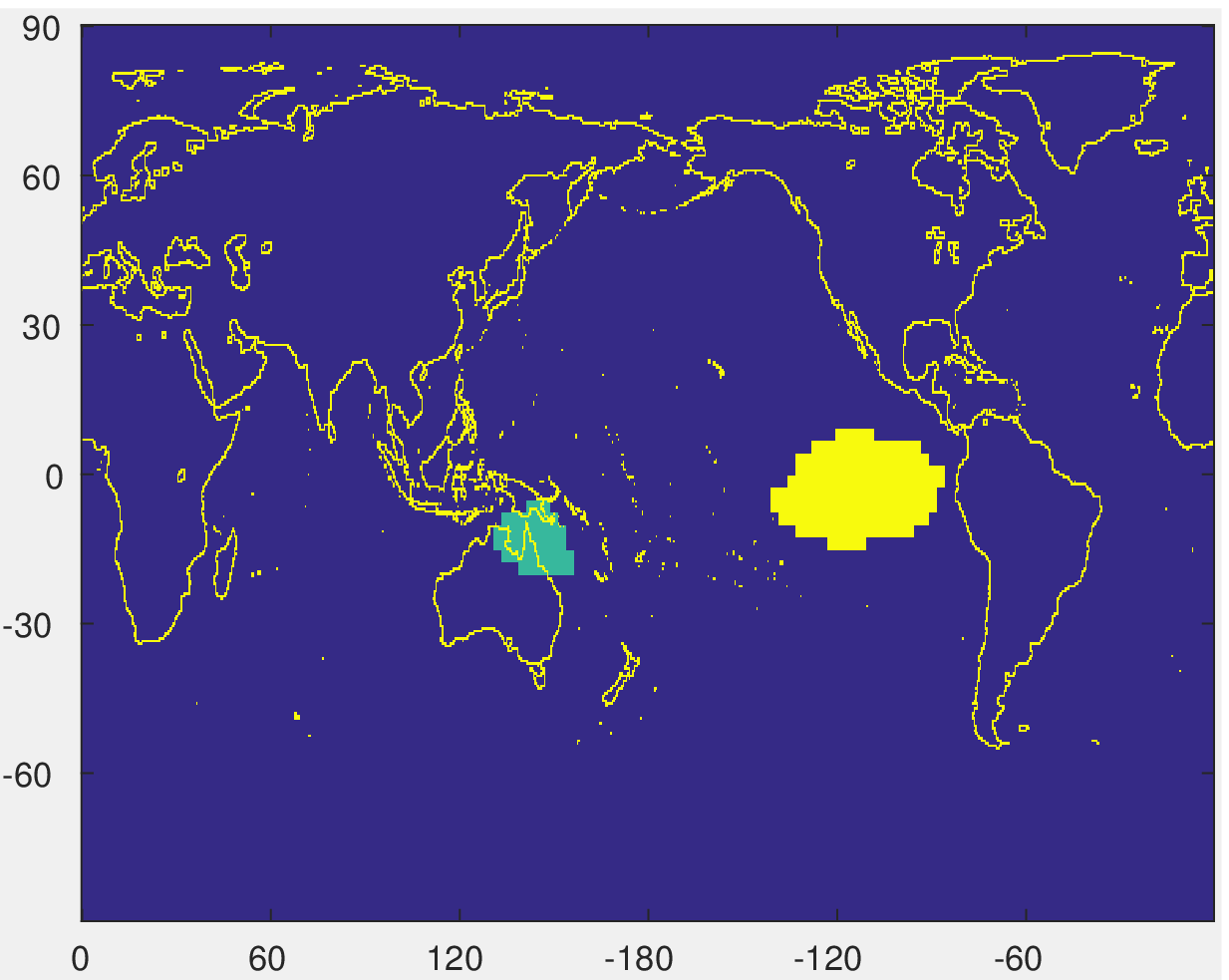}
		\centering
		\caption{Two ends of El-nino Southern Oscillation}
		\label{fig:ENSOReg}
	\end{subfigure}
	\begin{subfigure}[t]{0.7\textwidth}
		\includegraphics[scale=0.4]{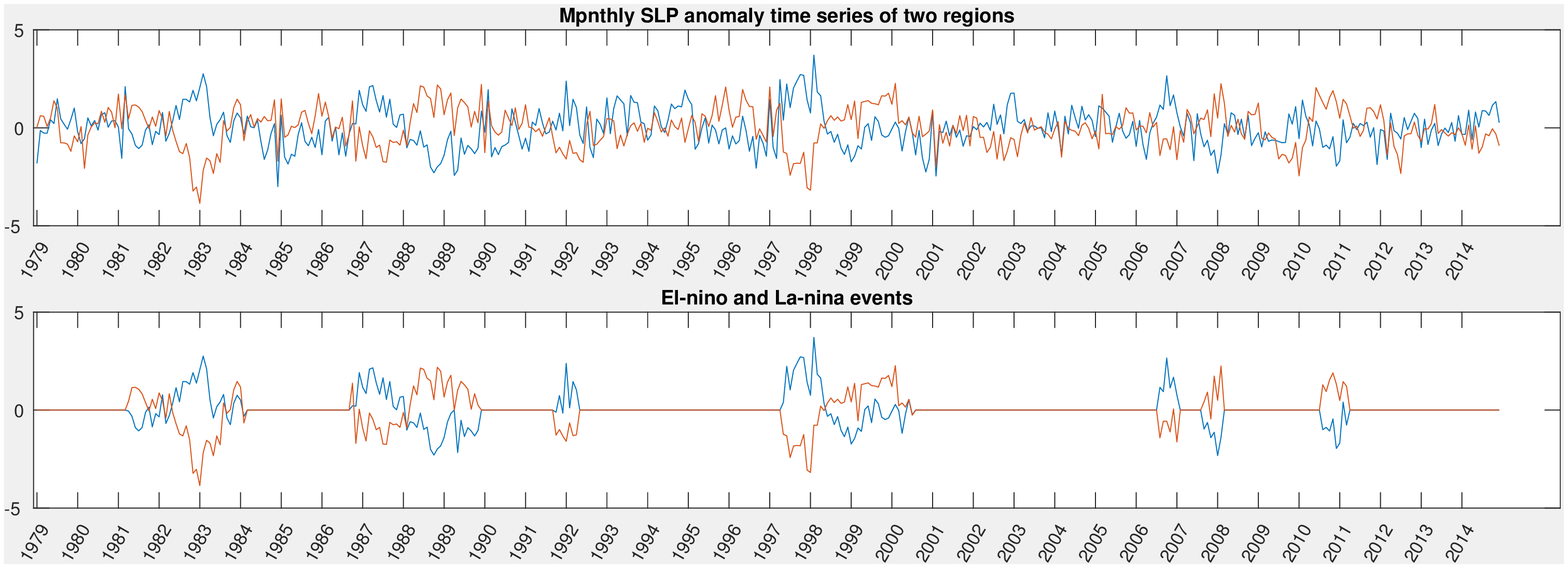}
		\centering
		\caption{Area-averaged Sea Level Pressure (SLP) monthly time series of two regions}
		\label{fig:ENSOTs}
	\end{subfigure}
	\vspace{-0.5em}
	\caption{An example of a Sub-Interval Relationship from climate science}
	\vspace{-0.5em}
\end{figure*}

Most of the existing work on mining time series relationships assume the relation to be present for the entire duration of the two time series. The most prevalent work has been done in designing various similarity measures (e.g., euclidean distance, Pearson correlation, dynamic time warping) for  analyzing full-length time series~\cite{kawale2013graph,keogh2002exact,liao2005clustering}. Another part of the related work goes into devising various largest common subsequence (LCS) matching problems \cite{das1997finding,chen2007spade,faloutsos1994fast}. Other related works focus on all-pairs-similarity-search and motif discovery~\cite{yeh2016matrix,zhu2016matrix}.

However, many interesting relationships in real-world applications often are intermittent in nature, i.e., they are highly prominent only in certain sub-intervals of time and absent or occur feebly in rest of the sub-intervals. As a motivating example, consider a pair of monthly Sea Level Pressure anomaly time series during 1979-2014 in Figure~\ref{fig:ENSOTs} that are observed at two distinct regions on the globe in Figure~\ref{fig:ENSOReg}). The full-length correlation between the two time series is $-0.25$. However, as shown in the lower panel of Figure~\ref{fig:ENSOTs}, there exists multiple sub-intervals where the correlation between the two time series is stronger than $-0.7$. As we discuss later in Section~\ref{Sec:DomInsights}, this example is the outcome of a well-known climate phenomena called ENSO (El Nino Southern Oscillation) \cite{glantz2001currents}, that is characterized by negative correlations between the surface temperatures observed near Australia and Pacific Ocean \cite{glantz2001currents} and is known for impacting various weather events such as floods, droughts, and forest fires \cite{siegert2001increased,ward2014strong}. The sub-intervals shown in the lower panel correspond to the two extreme phases, `El-Nino' and `La-nina', of ENSO, when its impact on global climate is amplified. Similar examples are also known to exist in other domains such as neuroscience, \cite{atluri2014discovering} and stock market data \cite{li2016efficient}.  


Inspired by such real-world examples, we   formally  define the notion of   SIR and devise necessary interestingness measures to characterize them.  We propose a novel and efficient approach called \textit{Partitioned Dynamic Programming}(PDP) to find the most interesting SIR in a given pair of time series.   We show that our approach is guaranteed to find the optimal solution and has time complexity that is practically linear in the length of the time series. 

\section{Definitions and Problem Formulation}\label{Sec:Def}
\begin{definition}
	A \textit{Sub-Interval Relationship (SIR)} between two time series $T_1$ and $T_2$ refers to the set $S$ of non-overlapping time intervals $\{[s_1,e_1],[s_2,e_2],...,[s_n,e_n]\}$ such that every interval in $S,$
	\begin{itemize}\setlength\itemsep{-0.5em}
		\item captures strong relationship between $T_1$ and $T_2$, i.e. $rel[s,e] \geq \tau$ $\forall [s,e] \in S.$
		\item is of length at least  $l_{min}$, i.e. $l_{se} \geq l_{min}\forall [s,e] \in S.$
	\end{itemize}
	where $\tau$ and $l_{min}$ are user-specified thresholds, and $rel()$ refers to a similarity measure.      
\end{definition}  
The choice of thresholds depends on the type of SIRs that are of interest to a user. For instance, setting higher $l_{min}$  and lower $\tau$ results in SIRs with longer intervals of mild relationships and vice-versa. 

\textbf{Problem Formulation:} Intuitively, an SIR is likely to be more reliable if the set $S$ of selected intervals covers a larger fraction of the timestamps. Therefore, we  measure interestingness of an SIR as the \textbf{\textit{sum-length}} ($SL$), which is equal to sum of lengths of all the selected sub-intervals. Then the problem require us to find the set of long and strong time-intervals with maximum sum-length. Formally, for a given pair of time series $(T_1,T_2)$, the goal is to determine the \textbf{optimal SIR} that has the largest sum-length. 

%

\vspace{-0.5em}
\section{Methodology}\label{Sec:Method}
Our problem formulation can potentially be solved by two approaches: (i) a classical approach based on the dynamic programming, and (ii) our proposed approach -- Partitioned Dynamic Programming, that is an extension of the classical dynamic programming. 
\vspace{-0.5em}
\subsection{Classical Approach: Dynamic Programming}\label{Sec:DP}
The problem of finding the optimal set can be treated as a classical DP problem of weighted interval scheduling \cite{kleinberg2005algorithm} where the goal is to determine a schedule of jobs such that no two jobs conflict in time and the total sum of the weights of all the selected jobs is maximized. In our problem, we can treat every time-interval that meets the minimum strength and length criteria as a job and the interval length as the weight of the job. We could then use DP to find the set of intervals with maximum possible sum-length. 

\textbf{Time Complexity}: It can be shown that both the average-case and worst-case time complexity of DP approach is $O(N^2)$ in time, where $N$ is the length of the time series. 



\subsection{Proposed Approach (Partitioned Dynamic Programming)}\label{sec:PDP}
DP's $O(N^2)$ complexity poses a serious challenge for long time series commonly found in climate and neuroscience domains. A potential approach to reduce computational cost could be to partition the original problem into multiple sub-problems of constant sizes and solving each of them independently using DP. The optimal set to the original problem could then be obtained by taking union of the optimal sets obtained for all sub-problems. The computational cost of   would then depend on the sizes of the different sub-problems. If the size of each sub-problem is smaller than a constant $k$, then the computational cost would be $O(\frac{N}{k}*k^2) = O(N)$, which would be faster than DP by an order of $N$. However, a key challenge in this approach is to partition the problem prudently such that no interesting interval gets fragmented across the two partitions, otherwise it could potentially get lost if its fragmented parts are not sufficiently long or strong enough to meet the user-specified thresholds.

To this end,  we propose a novel approach called \textit{Partitioned Dynamic Programming} (PDP) that is significantly efficient than Dynamic Programming (DP) and is guaranteed to find the optimal set. PDP follows the above idea and breaks the original problem into multiple sub-problems such that each one of them can be solved by using DP independently.
The key step in PDP is to identify safe \emph{points of partition}, where the problem could be partitioned without compromising on the optimality of the solution. However,   PDP is applicable only to those relationship measures that satisfy the following three properties: 

\noindent \textbf{Property 1:} The relationship measure could be computed over a single timestamp. \\
\textbf{Property 2:} If $rel[s,e]$ is known, then $rel[s,e+1]$ and $rel[s-1,e]$ could be computed in constant time. \\
\textbf{Property 3:} For a given pair of time series, let $[s,m]$ and $[m+1,e]$ be two adjacent time-intervals, $\alpha = min(rel[s,m]$, $ rel[m+1,e])$, and $\beta = max(rel[s,m]$, $ rel[m+1,e])$, then $\alpha \leq rel[s,e]$ $\leq$ $\beta$. \\
The above three properties are satisfied by various measures that we discuss in more detail in section~\ref{sec:Meas}.

From Property 3, it follows that an interval $[s,e]$ formed by   union of two adjacent weak intervals $[s,m]$ and $[m+1,e]$ could never be strong. Thus, a timestamp $t$ can be considered as a `point of partition' if:
\vspace{-0.5em}
\begin{enumerate}  \setlength\itemsep{0em}
	\item none of the intervals ending at $t-1$ are strong, i.e. $rel[s,t-1]< \tau \forall s \in [1,t-1]$. We refer to this condition as \textbf{left-weakness} condition.
	\item none of the intervals beginning from $t$ are \emph{strong}, i.e. $rel[t,e] < \tau \forall e \in [t,L]$. We refer to this condition as \textbf{right-weakness} condition. 
\end{enumerate}
\vspace{-0.5em}
The two conditions above ensure that all the intervals ending at $t-1$ or beginning from $t$ are weak, therefore no strong interval that subsumes $t$ could possibly exist.  Therefore, no interesting interval would be in danger of getting fragmented, if the problem is partitioned at $t$. Following this idea, we propose a partitioning scheme to find the points of partition before applying dynamic programming module to each of the partitions. 

PDP comprises of three major steps: In step 1, we find all timestamps $t$ such that they satisfy the left-weakness property. In step 2, we identify all the timestamps $t$ that satisfy the right-weakness property. Finally, in step 3, all the timestamps that satisfy both left-weakness and right-weakness are considered as the points of partition. The original problem is then partitioned at the obtained points of partition and the resulting sub-problems are solved independently using the DP module described in Section~\ref{Sec:DP}. 
\subsubsection{Finding timestamps with left-weakness:} To find timestamps with left-weakness, we perform a left-to-right scan of timestamps as follows. We begin our scan from the leftmost timestamp to find the first timestamp s such that $rel[s]\geq \tau$. We next show that all the timestamps $\{2,...,s\} $ will satisfy left-weakness using the following lemma.
\begin{lemma}\label{lem1}
	Consider a timestamp $t$ that satisfies left-weakness. If $rel[t] < \tau$, then $t+1$ would also satisfy left-weakness. 
\end{lemma}
\vspace{-0.5em}
Since there are no timestamps to the left of first timestamp, it trivially satisfies left-weakness. By recursively applying Lemma~\ref{lem1}, to timestamps $\{2,3,...,s\}$, we get each of them to satisfy left-weakness. 

We then continue our scan beyond $s$ to find the first timestamp $e$ such that $[s,e]$ is weak, i.e. $rel[s,e] < \tau$ (lines 15-21). This also means that for all the timestamps $m$ $\in$ $[s,e]$, interval $[s,m-1]$ is strong, and therefore $m$ violates left-weakness. We next claim that timestamp $e+1$ satisfies the left-weakness based on following lemma.
\begin{lemma}\label{lem2}
	Consider a set of timestamps $S =\{s,s+1,...,e-1,e\}$ such that $rel[s,m] \geq \tau \forall m \in [s,e-1]$, while $rel[s,e] < \tau$. If $s$ satisfies left-weakness, then timestamp $e+1$ would also satisfy left-weakness.  
\end{lemma}
We further continue our scan and repeat the above steps to find all the timestamps that satisfy left-weakness. In summary, the above procedure essentially finds streaks of points that satisfy or violate left- weakness in a single scan. Similar procedure could be followed to find timestamps that satisfy right-weakness except that the scan would proceed leftwards starting from the rightmost timestamp. 

\subsubsection{Time Complexity}\label{Sec:PDPcomplx}
There are three major steps in PDP. In step 1, we scan all the timestamps to find the ones that satisfy left-weakness. Notice that each timestamp is visited exactly once in the scan and under the assumption of Property 2, the computation of $rel[s,t]$ for every $t$ takes constant time. Therefore, the time complexity of step 1 is $O(N)$. Similarly, the complexity of Step 2 (to find points satisfy right-weakness) will also be $O(N)$. Step 3 solves the problem for each partition using standard DP. The total time complexity of PDP is therefore $O(N)+O(Nk)$, where k is the length of largest partition. For most cases, the threshold $\tau$ on relationship strength in each sub-interval is set to very high value which typically prohibits any partition to grow beyond a constant $k$ that is invariant to the length of time series. As a result, the time complexity of PDP turns out to be $O(N)$.



\subsection{Measures That Qualify For PDP}\label{sec:Meas}
Following popular relationship measures  satisfy all the three properties as discussed above, \\
1)  \textbf{Mean Square Error (MSE)} is calculated for a  given  pair of time series $(X,Y)$ as, $MSE[s,e] = \frac{(\sum\limits_{t=s}^e X[t] - Y[t])^2}{l_{se}}$ \\
2) \textbf{Average Product (AP)}    is given by $\frac{\sum\limits_{t=s}^{e}X[t]*Y[t]}{l_{se}+1}$.

\section{Results and Evaluation}\label{Sec:Eval}

\subsection{Data and Preprocessing}\label{SubSec:Data}
\textbf{Global Sea Level Pressure (SLP) Data:} We used monthly SLP dataset provided by NCEP/National Center for Atmospheric Research (NCAR) Reanalysis Project \cite{kistler2001ncep} which is available from 1979-2014 (36 years x 12 = 432 timestamps) at a spatial resolution of 2.5 $\times$ 2.5 degree (10512 grid points, also referred to as locations). 
In total, we sampled 14837 such pairs of regions from entire globe, whose time series have a full-length correlation weaker than 0.25 in magnitude. 

\subsection{Experimental Setup}\label{SubSec:ExptSetup}
\textbf{Similarity Measure:} Negative correlations in climate science have been widely studied in climate science. Hence, we adopt  \textit{negative Average Product} (nAP), which is exactly equal to the negative of $AP$ measure.



\textbf{Choice of parameters}: Our problem formulation requires inputs for two parameters: $l_{min}$, the minimum length and $\tau$, the minimum strength of relationship in every sub-interval of SIR. In climate science, a physical phenomenon typically shows up as a strong signal that lasts for at least six months, hence we chose $l_{min} = 6$ for SLP data.  The other parameter $\tau$ was set to a high value of 1. 




\vspace{-0.5em}
\subsection{Computational Evaluation}\label{sec:EvalComp}
\begin{figure}[t]
	\centering
	\includegraphics[scale=0.4]{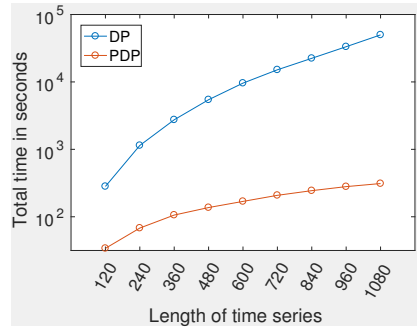}
	\caption{Computing time (Y-axis) of DP and PDP for different lengths of time series (X-axis).} 
	\label{fig:PDPVsDPDiffLen}
	\vspace{-1em}
\end{figure}

We evaluate PDP against DP based on their computational costs and their scalability on datasets with long time series. To generate datasets of different sizes, we used global SLP data simulated by GFDL-CM3, a coupled physical climate model that provides simulations for entire $20^{th}$ century from which we obtained nine time-windows of different sizes, each starting from 1901 and ending at 1910, 1920,...,1990 and for each time window, we obtained a set of 14837 pairs of time series.  Figure~\ref{fig:PDPVsDPDiffLen} shows the total computational time taken by DP and PDP to find SIRs in all the datasets. As can be seen, the computing time of DP (blue curve) follows a quadratic function of the length of time series, whereas the same for PDP (red curve) increases linearly with the length of time series. This is no surprise because the time complexity of PDP is $O(kN)$ in time, where the constant $k$ is governed by the length of largest partition. Typically, the size of sub-intervals exhibiting strong relationships does not exceed beyond a constant size and therefore, is independent of the length of time series, which makes PDP linear in run-time.

\begin{figure}[!h]
	\centering
	\includegraphics[scale=0.4]{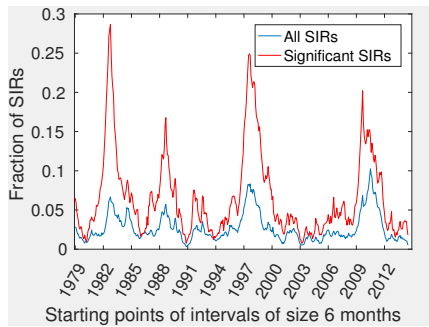}
	\centering
	\caption{For every interval of size 6 months, the plot indicates the proportion of i) 1044 significant SIRs (red curve) and ii) all 14837 candidate SIRs (blue curve) that included given interval.}\label{fig:EventSLP}
	\vspace{-1.5em}
\end{figure}

\vspace{-0.7em}
\subsection{Applications and Domain Insights}\label{Sec:DomInsights}
\textbf{Finding Anomalous Intervals}: A potential application of this work could be to detect anomalous time intervals that experience unusually high number of relationships. Specifically, for every interval $[s,e]$, one can obtain a score that indicates the proportion of candidate pairs that were 'active' during entire $[s,e]$. Intervals included in unusually high number of SIRs could potentially indicate the occurrence of a special event. Applying this idea to SIRs of SLP dataset, we obtained the scores for all possible intervals of size 6 months as shown in Figure~\ref{fig:EventSLP}. It can be seen that the scores are anomalously high for the intervals 1982 Sept-83 Mar, 1988 Sept -89 Mar, 1997 Aug -98 Feb, and 2009 Sept -10 Mar. All of the above intervals are known to have experienced the strongest el-nino and la-nina events since 1979 \cite{ensoyears}. During these events, climate behaves quite differently compared to the general climatology. New wave patterns emerge that synchronize regions with each other that are otherwise unrelated to each other. 

\vspace{-0.7em}
\vspace{-0.5em}
\section{Conclusion}\label{Sec:Conc}

In this paper, we defined a notion of sub-interval relationship to capture interactions between two time series that are intermittent in nature and are prominent only in certain sub-intervals of time. We proposed a fast-optimal algorithm to find most interesting SIR in a pair of time series. 
We further demonstrated the utility of  SIR in climate science applications and obtain useful domain insights. 

\bibliography{refs}
\bibliographystyle{icml2019}

\section{Lemma Proofs}\label{Sec:Proofs}

\noindent \textbf{Proof Lemma 1}: Consider two adjacent intervals $[a,t-1]$ and $[t]$ for any $a \in [1,t-1]$. Since $t$ satisfies left-weakness, $rel[a,t-1] < \tau$. Also, $rel[t] < \tau$, therefore from Property 3, $rel[a,t] < \tau \forall a \in [1,t]$. Thus, $t+1$ satisfies left-weakness. 



\noindent \textbf{Proof Lemma 2}: Following the definition of left-weakness, it would suffice to show that $rel[t,e] < \tau, \forall t \in [1,e]$.
We prove this in two parts: In first part, we show that $rel[t,e] < \tau, \forall t \in [s,e]$, while in second, we show that $rel[t,e]< \tau,  \forall t \in [1,s-1]$.

\textbf{Part 1:} For any $m \in [s,e-1]$, consider two adjacent intervals $[s,m]$ and $[m+1,e]$. Let $\alpha = min(rel[s,m]$, $ rel[m+1,e])$ and $\beta = max(rel[s,m]$, $rel[m+1,e])$. Then from Property 3, we get $\alpha \leq rel[s,e] \leq \beta$.
Since $rel[s,e] < \tau$,  we have $\alpha < \tau$.
Since $rel[s,m] \geq \tau$, thus ,we get $\alpha = rel[m+1,e]$ and,
\begin{equation}\label{Eq3}
rel[m+1,e]< \tau, \forall m \in [s,e]
\end{equation}
\textbf{Part 2:} We know that $s$ satisfies left-weakness. Therefore, by definition,
$rel[t,s-1]< \tau, \forall t \in [1,s-1]$. Also we have $rel[s,e] < \tau$. Thus, putting $m = s-1$ in Property 3, we get
\begin{equation}\label{Eq4}
rel[t,s]< \tau \forall t \in [1,s-1] 
\end{equation}
Together from Eqs~(\ref{Eq3}) and (\ref{Eq4}), we get 
$$ rel[t,e]< \tau \hspace{1em} \forall t \in [1,e] $$

\end{document}